\documentclass[10pt,twocolumn,letterpaper]{article}

\usepackage{cvpr}              % To produce the CAMERA-READY version
\usepackage{graphicx}
\usepackage{amsmath}
\usepackage{amssymb}
\usepackage{booktabs}
\usepackage[table,xcdraw]{xcolor}

\usepackage{times}
\usepackage{epsfig}
\usepackage{multirow}
\usepackage{url}
\usepackage{bm}
\usepackage{color}
\usepackage{booktabs}
\usepackage{caption}

\newcommand{\mypar}[1]{\noindent\textbf{#1}}

\usepackage[pagebackref,breaklinks,colorlinks]{hyperref}

\usepackage[capitalize]{cleveref}
\crefname{section}{Sec.}{Secs.}
\Crefname{section}{Section}{Sections}
\Crefname{table}{Table}{Tables}
\crefname{table}{Tab.}{Tabs.}

\def\cvprPaperID{xxx} % *** Enter the CVPR Paper ID here
\def\confName{CVPR}
\def\confYear{2022}

\begin{document}

\title{PVSeRF: Joint Pixel-, Voxel- and Surface-Aligned Radiance Field for Single-Image Novel View Synthesis}

\author{Xianggang Yu$^{1}$, Jiapeng Tang$^{2}$, Yipeng Qin$^{3}$, Chenghong Li$^{1}$\\
Linchao Bao$^{4}$, Xiaoguang Han$^{1}$, Shuguang Cui$^{1}$\\ 
$^1$The Chinese University of Hong Kong, Shenzhen \quad $^2$Technical University of Munich \\
$^3$Cardiff University \quad $^4$Tencent AI Lab}

\maketitle

% Main Content
\begin{abstract}
	We present PVSeRF, a learning framework that reconstructs neural radiance fields from single-view RGB images, for novel view synthesis. 
	Previous solutions, such as pixelNeRF~\cite{yu2021pixelnerf}, rely only on pixel-aligned features and suffer from feature ambiguity issues.
	As a result, they struggle with the disentanglement of geometry and appearance, leading to implausible geometries and blurry results. 
	To address this challenge, we propose to incorporate explicit geometry reasoning and combine it with pixel-aligned features for radiance field prediction. 
	Specifically, in addition to pixel-aligned features, we further constrain the radiance field learning to be conditioned on i) voxel-aligned features learned from a coarse volumetric grid and ii) fine surface-aligned features extracted from a regressed point cloud.
	We show that the introduction of such geometry-aware features helps to achieve a better disentanglement between appearance and geometry,~\ie recovering more accurate geometries and synthesizing higher quality images of novel views.
	Extensive experiments against state-of-the-art methods on ShapeNet benchmarks demonstrate the superiority of our approach for single-image novel view synthesis. 

\end{abstract}
\section{Introduction}
\label{sec:introduction}

\begin{figure}[t]
	\centering
	\includegraphics[width=1.\linewidth]{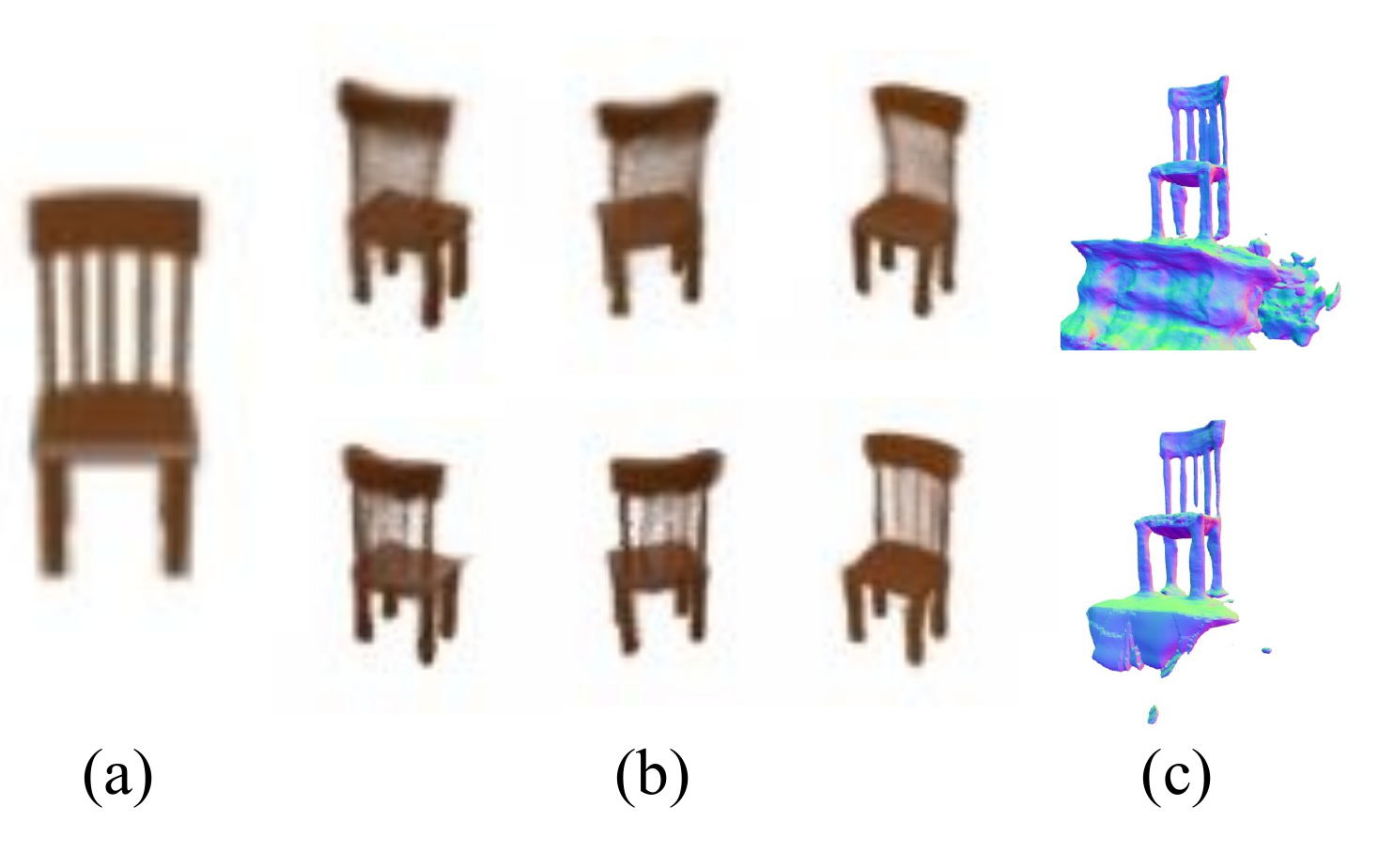}
	\caption{\textbf{Novel view synthesis from a single image.} 
	     (a) Input image. (b) Novel view synthesis results: pixelNeRF~\cite{yu2021pixelnerf} (top) and Ours (bottom). (c) Surface meshes extracted from predicted radiance fields: pixelNeRF~\cite{yu2021pixelnerf} (top) and Ours (bottom). By augmenting the 2D pixel-aligned features with complementary 3D geometric features for radiance field prediction, we can synthesize higher quality of novel views. A by-product of our approach is a cleaner implicit surface mesh, due to the introduction of explicit geometric features.
	}
	
	%We present a novel approach that learns to reconstruct neural radiance fields for single-image novel view synthesis. (a) Input image. (b) Novel view synthesis results: PixelNeRF~\cite{yu2021pixelnerf} (top) and Ours (bottom). (c) Meshes extracted from predicted radiance fields: PixelNeRF~\cite{yu2021pixelnerf} (top) and Ours (bottom). By constraining radiance field learning on geometry-aware features, a by-product of our method is acquiring a cleaner implicit geometry. In contrast, the underlying geometry of PixelNeRF~\cite{yu2021pixelnerf} bears much noise due to the crucial feature ambiguity issue.
	\label{fig:teaser}
	\vspace{-5mm}
\end{figure}

Novel view synthesis is a long-standing problem in computer vision and graphics, which plays a crucial role in various practical applications, including gaming, movie production, and virtual/augment reality. 
Recently, it has made great strides thanks to the advances in differentiable neural rendering~\cite{niemeyer2020differentiable, yariv2020multiview}, especially the neural radiance fields (NeRF)~\cite{mildenhall2020nerf} that simplifies novel view synthesis to an optimization problem over a dense set of ground truth views.

Although achieving impressive results, the vanilla NeRF suffers from several limitations: i) the dense views it strictly requires are not always available; ii) it is slow in inference due to the long optimization process; iii) each NeRF is dedicated to a specific scene and cannot be generalized to new ones. 

\begin{figure}[t] %htb
	\centering
	\includegraphics[width=1.\linewidth]{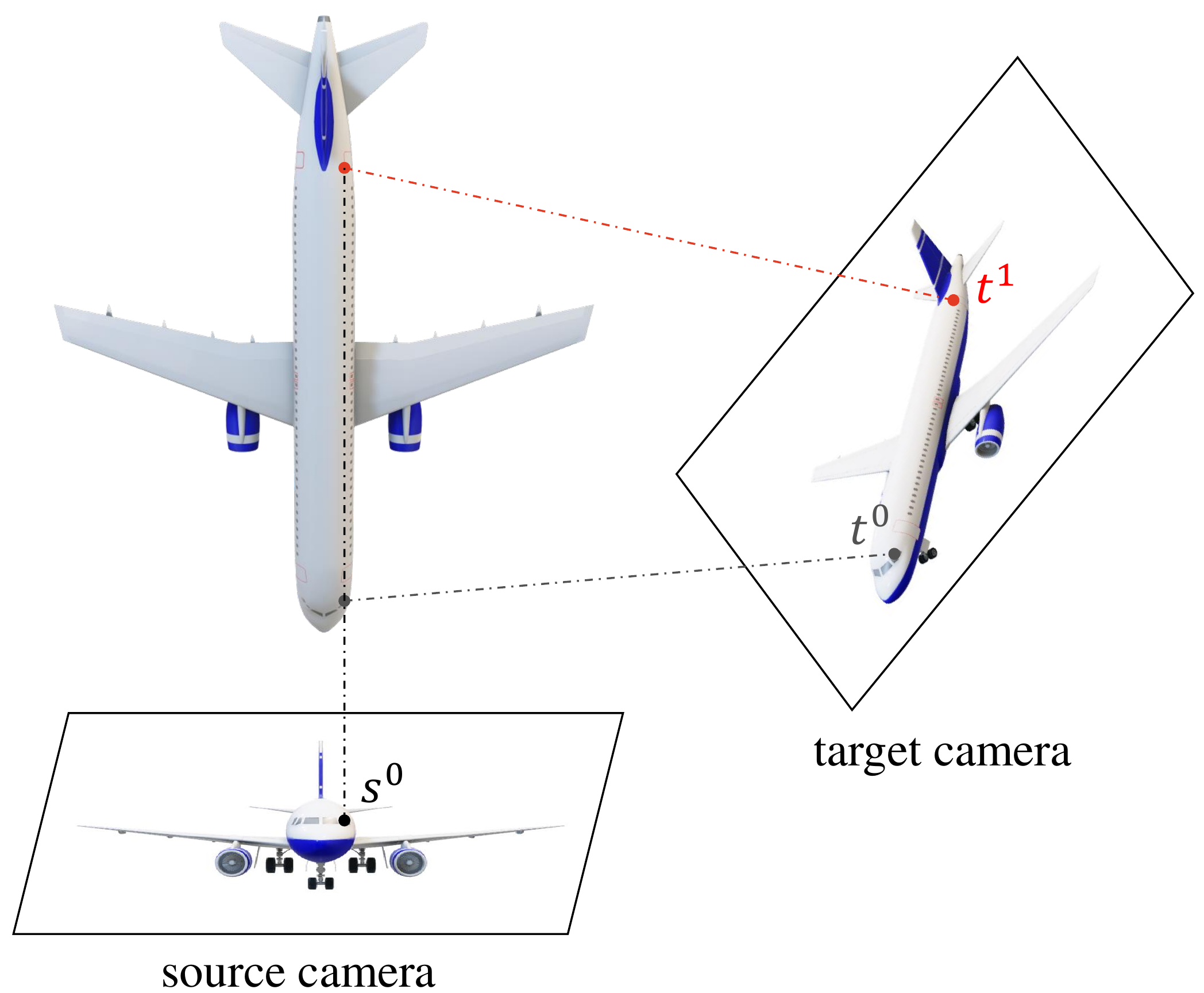}
	\caption{\textbf{Illustration of the feature ambiguity issue.} The feature ambiguity issues that originates from the {\it many-to-one} mapping between queries and their corresponding pixel-aligned features. Two rays shot from points $t^{0}$ and $t^{1}$ on target camera intersect the same ray shot from source camera. After the pixel-aligned process, the two different intersecting points will be projected to the same image coordinate $s^{0}$ on image-plane, obtaining the same pixelwise feature.}
	\label{fig:visibility_issue}
	\vspace{-3mm}
\end{figure}
To address these issues, follow-up works such as pixelNeRF~\cite{yu2021pixelnerf}, IBRNet~\cite{wang2021ibrnet}, and GRF~\cite{trevithick2020grf}, proposed to predict neural radiance fields in a feed-forward manner. 
Taking pixelNeRF as an example, it tackles the shortcomings of NeRF by extending its network to be conditioned on scene priors learnt by a convolutional image encoder.
These scene priors are represented by spatial feature maps that allow the mapping from a pair of query spatial point and viewing direction to their corresponding pixel-aligned features.
In pixelNeRF, such a mapping is implemented by standard camera projection and bilinear interpolation.
During inference, the scene priors are obtained via a forward pass through the image encoder and thus allow fast novel view synthesis from a single input view of diverse scenes.
Although effective, pixelNeRF suffers from feature ambiguity issues that originates from the {\it many-to-one} mapping between queries and their corresponding pixel-aligned features.
In other words, pixelNeRF naively assigns the same pixel-aligned features to different points in some novel view as long as these points overlap with each other in the input view, which can cause confusion (Fig.~\ref{fig:visibility_issue}).

To clarify such ambiguity issues, we propose to incorporate explicit geometry reasoning and combine it with pixel-aligned features for radiance field prediction.
Specifically, we leverage the recent success in single-view 3D reconstruction~\cite{choy20163d, fan2017point, groueix2018atlasnet, tang2019skeleton, pan2019deep, mescheder2019occupancy, chen2019learning, tang2020skeletonnet} and inject rich geometry information into radiance field prediction by incorporating geometry-aware features of two shape representations: i) voxel-aligned features learned from a coarse volumetric grid and ii) fine surface-aligned features extracted from a regressed point cloud.
Intuitively, such geometry-aware features augment pixel-aligned features with additional ``dimensions'', thereby allowing previously ambiguous points to be separable.
Furthermore, by constraining the radiance field learning on these geometry-aware features, our method not just synthesize higher quality images of novel views, but also recover more accurate underlying geometries in radiance field, as witnessed in~\cref{fig:teaser}.

Our main contributions include:
\begin{itemize}

	\item We propose a novel approach of learning neural radiance fields from single-view images jointly conditioned on pixel-, voxel-, surface-aligned features.
	\item We design an efficient way to alleviate the feature ambiguity issue of solely pixel-aligned features by incorporating explicit geometry reasoning via single-view 3D reconstruction.
	\item We propose a hybrid use of geometric features, including complementary coarse volumetric features and fine surface features. 
	
\end{itemize}

\section{Related Work}
\label{sec:related_work}

% \begin{figure*}[htb]
% 	\centering
% 	\includegraphics[width=1.\linewidth]{figures/PVSNeRF-pipeline_v3}
% 	\caption{\textbf{Overview of our PVSeRF framework.} Given a single input image $\mathbf{I}$, we first extract the spatial feature map $\mathbf{F_I}$ using a fully convolutional image encoder $E$, learn a volumetric grid $\mathbf{V}$ through volume generator $D_V$, and regress a sparse point set $\mathcal{S}$ of the object through a point set generator $G_\mathcal{S}$. From $\mathbf{V}$ and $\mathcal{S}$, we can learn geometric features $\mathbf{F_V}$ and $\mathbf{F}_\mathcal{S}$. Then, for a 3D location $x$ and a target view direction $\mathbf{d}$, we query pixel-, voxel-, and surface-aligned $\mathbf{f_I}$, $\mathbf{f_V}$, $\mathbf{f}_{\mathcal{S}}$ from $\mathbf{F_I}$, $\mathbf{F_V}$, and $\mathbf{F}_\mathcal{S}$ respectively. Next, the 3D location, view direction and all corresponding features passed into a MLP to predict density $\sigma$ and radiance $\mathbf{r}$. Lastly, the volume rendering is used to accumulate the radiance prediction of points in the same ray to compute the final color values.}
% 	\label{fig:pipeline}
% 	\vspace{-5mm}
% \end{figure*}

\begin{figure*}[htb]
	\centering
	\includegraphics[width=.9\linewidth]{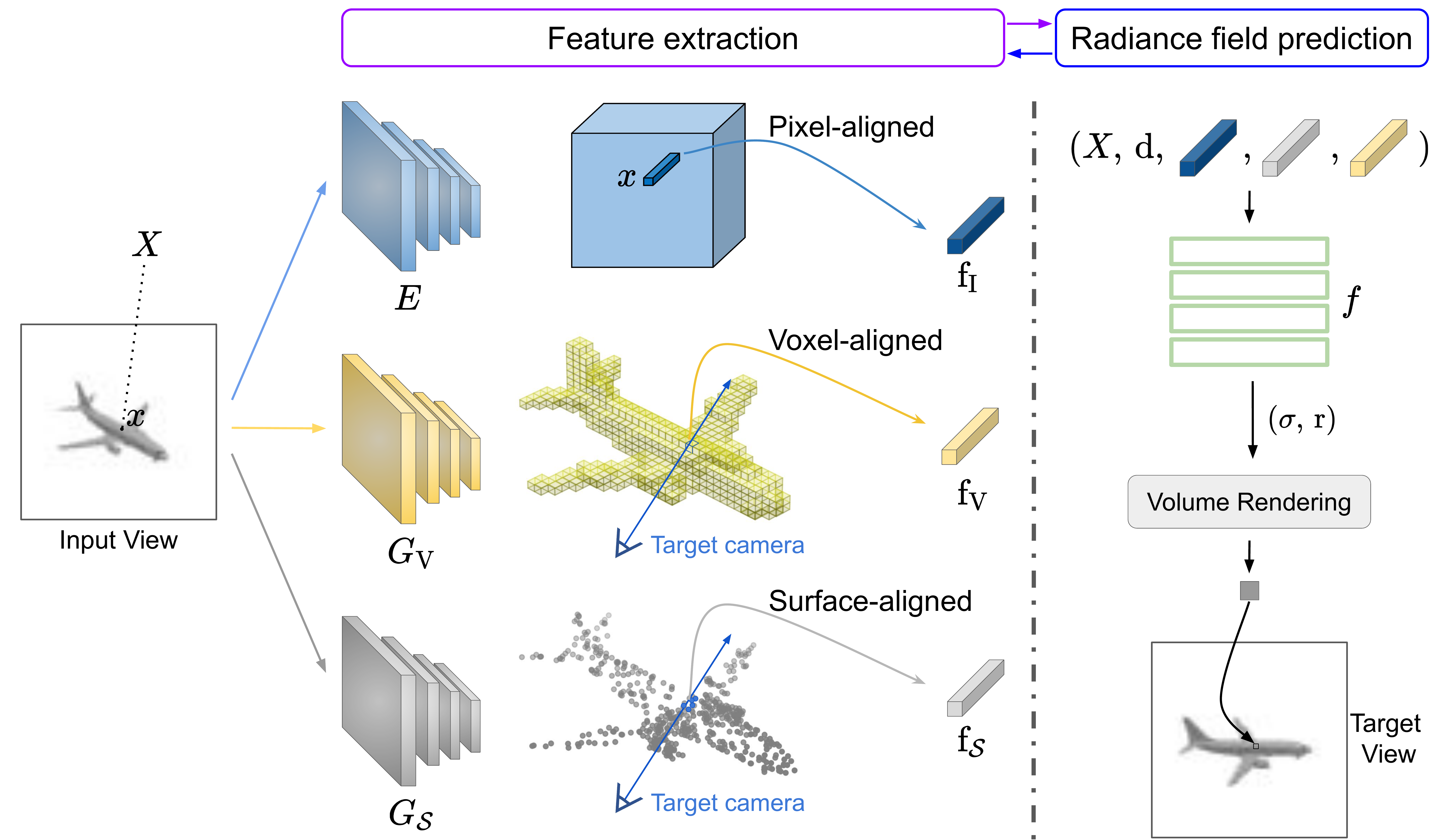}
	\caption{\textbf{Overview of our PVSeRF framework.} Given a single input image, we first 1) extract the spatial feature map using a fully convolutional image encoder $E$, 2) learn a volumetric grid through volume generator $G_\mathbf{V}$, and 3) regress a surface point set of the object through a point set generator $G_\mathcal{S}$. From volumetric grid and surface point set, we can learn voxel features and point-wise features. Then, for a 3D location $\mathbf{X}$ and a target view direction $\mathbf{d}$, we query pixel-, voxel-, and surface-aligned $\mathbf{f_I}$, $\mathbf{f_V}$, $\mathbf{f}_{\mathcal{S}}$ from spatial feature map, voxel features and point-wise features respectively. Next, the 3D location, view direction and all corresponding features are directed into a MLP to predict density $\sigma$ and radiance $\mathbf{r}$. Lastly, the volume rendering is used to accumulate the radiance prediction of points on the same ray to compute the final color values.}
	\label{fig:pipeline}
	\vspace{-5mm}
\end{figure*}

\subsection{Novel view synthesis and Neural radiance field}
The task of novel view synthesis aims to generate new views of a scene from single or a set of sparse views. There are various kinds of approaches dedicated to this problem. Traditional methods ~\cite{gortler1996lumigraph, davis2012unstructured, levoy1996light} choose to estimate light fields and then render novel views. Recent years, with the advance of deep neural networks (DNN), a plethora of models are designed to learn novel view synthesis in an end-to-end manner. Pioneering methods~\cite{park2017transformation, tatarchenko2016multi, zhou2016view} consider it as a image-to-image transformation problem and directly utilize 2D CNN to output novel views. These methods always cannot generate satisfactory results for viewpoints that are largely deviated from the given view. 

Later work explore 3D-aware image synthesis and solve the inverse rendering problem via neural networks~\cite{sitzmann2019deepvoxels, wiles2020synsin, NguyenPhuoc2019HoloGANUL, Geng2018WarpguidedGF, Kossaifi2018GAGANGG, NguyenPhuoc2018RenderNetAD, Zhu2018VisualON}. The common characteristic of this line of literature is that they recover the explicit or implicit 3D geometry and appearance properties first, then render novel views at desired camera viewpoints by means of differentiable rendering techniques~\cite{NguyenPhuoc2018RenderNetAD} or generative models. Among these work, various 3D representations are employed. DeepVoxels~\cite{sitzmann2019deepvoxels} represents 3D scene properties by low-resolution volumetric feature grid lifted from 2D feature maps. Wiles~\etal~\cite{wiles2020synsin} use 3D surface features that are learned from the point cloud unprojected from the estimated depth map of the input view. Other approaches~\cite{NguyenPhuoc2019HoloGANUL, Geng2018WarpguidedGF, Kossaifi2018GAGANGG, Zhu2018VisualON} learn implicit 3D embedding that can be used to generate novel views of the same scene using unsupervised learning techniques. 

Recently, witnessing the great success of neural radiance field (NeRF)~\cite{mildenhall2020nerf}, there has been an explosion of NeRF-based approaches for novel view synthesis\cite{MartinBrualla2021NeRFIT, Liu2020NeuralSV, Lindell2021AutoIntAI, Neff2021DONeRFTR, reiser2021kilonerf, yu2021plenoctrees, Tretschk20arxiv_NR-NeRF, park2021nerfies, chibane2021stereo, yu2021pixelnerf, wang2021ibrnet, trevithick2020grf, Chen2021MVSNeRFFG}. There are two divisions in the prevalence of NeRF: 1) the first track tries to train scene-specific model for generating novel views of the scene~\cite{MartinBrualla2021NeRFIT, Liu2020NeuralSV, Lindell2021AutoIntAI, Neff2021DONeRFTR, reiser2021kilonerf, yu2021plenoctrees, Tretschk20arxiv_NR-NeRF, park2021nerfies}. Specifically, they capture many diverse viewpoints of a scene, and optimizing a neural radiance field for that scene. Despite synthesizing high-fidelity novel views, these methods require  
longstanding optimization process and cannot generalize to new scenes. 2) the second track attempts to learn generalize neural radiance field across multiple scenes~\cite{chibane2021stereo, yu2021pixelnerf, wang2021ibrnet, trevithick2020grf, Chen2021MVSNeRFFG}. Among this, pixelNeRF~\cite{yu2021pixelnerf} is the most relevant method to ours, which learns the scene priors conditioned on the pixel-aligned features, and can switch to new scenes flexibly. Although other methods~\cite{chibane2021stereo, wang2021ibrnet, trevithick2020grf, Chen2021MVSNeRFFG} can also be applied to novel scene through a single forward pass, their methods are equipped to multiple input views, while we focus on the more challenging single-view input setting.

\subsection{Single-view 3D Object Reconstruction}
    Given a single image containing a object, 3D object reconstruction aims to recover the 3D geometry of the object. Traditional 3D reconstruction methods~\cite{hartley2000multiple, hirschmuller2007stereo, campbell2008using, tola2012efficient} need to find dense correspondence across multi-view at the first, followed by the depth fusion stage. Recently, due to the establishment of large-scale 3D model datasets, such as ShapeNet~\cite{chang2015shapenet} and ModelNet~\cite{wu20153d}, it is popular to reconstruct complete 3D shape from a single image by utilizing shape priors modeled by deep neural networks.
    It also achieved various degrees of success by designing 3D shape decoders tailored for different shape representations including voxel~\cite{choy20163d, han2017high}, point cloud~\cite{fan2017point, nguyen2019graphx}, mesh~\cite{groueix2018atlasnet, pan2019deep, tang2020skeletonnet}, and implicit field~\cite{chen2019learning, mescheder2019occupancy, park2019deepsdf, tang2020skeletonnet, peng2020convolutional, tang2021sa}.
    The voxel decoders~\cite{choy20163d, wu2016learning} take advantages of conventional 3D convolution operations to generate volumetric grids.  
    The point decoders~\cite{fan2017point, yang2018foldingnet} directly regress the coordinates of 3D points.
    The mesh decoders mainly approximate a target shape by performing template mesh deformation~\cite{groueix2018atlasnet, kato2018neural, pan2019deep, tang2019skeleton, tang2020skeletonnet}.
    The neural implicit functions~\cite{chen2019learning, mescheder2019occupancy, park2019deepsdf, tang2020skeletonnet, peng2020convolutional} represent 3D surfaces by continuous functions defined in 3D space. In this paper, we want to incorporate explicit geometry reasoning into the process of single-view novel view synthesis by marrying single-view 3D shape generators with a generic radiance field learning model.

\section{PVSeRF}
\label{sec:method}

\subsection{Overview}
   Given a single-view image $\mathbf{I}$ with its corresponding camera pose $\mathbf{R}$ and intrinsic shape parameter $\mathbf{K}$, our PVSeRF aims to learn a neural network for radiance field reconstruction:
   \begin{equation}
        \sigma, \mathbf{r} = \mathrm{PVSeRF} (\mathbf{X}, \mathbf{d}; \mathbf{I}, \mathbf{R}, \mathbf{K})
    \label{pvserf}
    \end{equation}
    where $\mathbf{X} \in \mathbb{R}^3$ represents a 3D location, $\mathbf{d} \in \mathbb{R}^2$ is a view direction, $\sigma$ is the volume density at $\mathbf{X}$, and $\mathbf{r}$ is the predicted radiance (RGB color) at $\mathbf{X}$ depending on the viewing direction $\mathbf{d}$.
    By accumulating the $\sigma$ and $\mathbf{r}$ of multiple points sampled on the ray defined by $\mathbf{X}$ and $\mathbf{d}$, we can obtain the color values of all pixels in a target view image $\mathbf{I_t}$ via differentiable rendering, thereby enabling novel view synthesis.

The distinct advantage of our PVSeRF is that it addresses the feature ambiguity issue of pixelNeRF~\cite{yu2021pixelnerf} by a novel geometric regularization using both voxel- and surface-aligned features.
As aforementioned, pixelNeRF's feature ambiguity issue stems from the fact that its network is solely conditioned on the 2D pixel-aligned features where multiple query 3D points are mapped to a single location.
To clarify this ambiguity, we propose to augment the 2D pixel-aligned features with complementary 3D geometric features for radiance field construction.
As Fig.~\ref{fig:pipeline} shows, in addition to the pixel-aligned features, our method incorporates i) voxel-aligned and ii) surface-aligned features into radiance field prediction.
Specifically,
\begin{itemize}
    \item We follow~\cite{yu2021pixelnerf} and extract the pixel-aligned features $\mathbf{f_I}$ of a query point $\mathbf{X}$ by projecting it with $\mathbf{R}$ and $\mathbf{K}$ to the 2D image coordinates $x$, indexing the multi-scale feature maps of an input image $\mathbf{I}$ extracted by a fully-convolutional image encoder $E$.
    \item We extract the voxel-aligned features $\mathbf{f_V}$ of a query point $\mathbf{X}$ by trilinearly interpolating $\mathbf{X}$ in a low-resolution volumetric feature $\mathbf{F_V}$ learnt from the input image $\mathbf{I}$ using a volume generator $G_\mathbf{V}$. Note that $\mathbf{f_V}$ only captures coarse geometry contexts of the scene due to the low-resolution nature of $\mathbf{F_V}$.
    \item To capture the geometric information on surface, we extract the fine-grained surface-aligned features $\mathbf{f}_\mathcal{S}$ of a query point $\mathbf{X}$ as the weighted sum of the associated features $\mathbf{F}_\mathcal{S}$ of its $K$ nearest neighbors in a point cloud $\mathcal{S}$, which is reconstructed from the input image $\mathbf{I}$ using a point set generator $G_{\mathcal{S}}$.
\end{itemize}
Thus, our PVSeRF is conditioned on $\mathbf{f_I}$, $\mathbf{f_V}$, and $\mathbf{f}_\mathcal{S}$ and can be reformulated as:
    \begin{equation}
        \sigma, \mathbf{r} = \mathrm{PVSeRF}(\mathbf{X}, \mathbf{d}; \mathbf{f_I} \oplus \mathbf{f_V} \oplus \mathbf{f}_\mathcal{S})
    \label{radiance}
    \end{equation}
where $\oplus$ denotes a concatenation operation. Thanks to the incorporation of $\mathbf{f_V}$ and $\mathbf{f}_\mathcal{S}$, the previously ambiguous points that share the same $\mathbf{f_I}$ are now separable by the concatenation $\mathbf{f_I} \oplus \mathbf{f_V} \oplus \mathbf{f}_\mathcal{S}$.  
We present more details about each component of our method as follows.

 \subsection{Feature Extraction}
   
\noindent \textbf{Pixel-aligned Features}
    Following pixelNeRF~\cite{yu2021pixelnerf}, we also use pixel-aligned features that contain fine-grained details about the scene's geometry and appearance properties to learn neural radiance fields.
    Given an input image $\mathbf{I} \in \mathbb{R}^{H \times W \times 3}$, we employ a fully-convolutional image encoder $E$ implemented by ResNet-34~\cite{he2016deep} to extract its multi-scale feature maps $ \{\mathbf{F_I^0}, \mathbf{F_I^1}, \mathbf{F_I^2}, \mathbf{F_I^3}\}$, which are the intermediate features at 'conv1', 'layer1', 'layer2', and 'layer3' of ResNet-34 but upsampled to the size of the input image $\mathbf{I}$.
    Then, we acquire the pixel-aligned feature vector $\mathbf{f_I}$ of a query 3D point $\mathbf{X}$ by projecting $\mathbf{X}$ to the 2D image coordinates $x$, and bilinearly interpolating the feature maps concatenated by $ \{\mathbf{F_I^0}, \mathbf{F_I^1}, \mathbf{F_I^2}, \mathbf{F_I^3}\}$ through $\mathcal{B}$:
     \begin{equation}
         \mathbf{f_I} = \mathcal{B}(\mathbf{F_I^0} \oplus \mathbf{F_I^1} \oplus \mathbf{F_I^2} \oplus \mathbf{F_I^3}, \mathbf{K}\mathbf{R}\mathbf{X})
         \label{eq:pixel_aligned_feature}
     \end{equation}
     where $\oplus$ represents feature concatenation.
     However, $\mathbf{KR}$ may project multiple 3D points $\mathbf{X}$ along $\mathbf{R}$ to a single position on the 2D image coordinates, leading to ambiguous $\mathbf{f_I}$ and blurry synthesized novel views. 
     To clarify such ambiguity, we propose to augment $\mathbf{f_I}$ with complementary geometric features, including both coarse voxel-aligned features learned from a volumetric grid, and fine surface-aligned features extracted from a regressed point cloud.
  
\vspace{2mm}  
\noindent  \textbf{Voxel-aligned Features}
    We compute the voxel-aligned feature $\mathbf{f_V}$ with respect to $\mathbf{X}$ as follows. 
    First, we reconstruct a volumetric feature grid $\mathbf{F_V} \in \mathbb{R}^{32 \times 32 \times 32 \times C}$ from the input image $\mathbf{I}$ using a volume generator consisting of a VGG-16~\cite{simonyan2014very} image encoder and a 3D CNN decoder. 
    Then, we have:
    \begin{equation}
        \mathbf{f_V} = \mathcal{T}(\mathbf{F_V}, \Omega(\mathbf{X}))
    \end{equation}
    where $\mathcal{T}$ is a multi-scale trilinear interpolation inspired by GeoPiFu~\cite{he2020geo} and IFNet~\cite{chibane2020implicit}, $\Omega(\mathbf{X})$ is a point set around $\mathbf{X}$:
    \begin{equation}
        \Omega(\mathbf{X}) = \{ \mathbf{X} + s \cdot \mathbf{n} | \mathbf{n} = (1,0,0), (0,1,0), (0,0,1), ...\}
    \end{equation}
    where $s \in \mathbb{R}$ is the step length, $\mathbf{n} \in \mathbb{R}^3$ represents the unit vectors defined along the three axes in a Cartesian coordinate system.
    Intuitively, $\mathbf{f_V}$ is a concatenation of all queried feature vectors at points in $\Omega(\mathbf{X})$ that are trilinearly interpolated from $\mathbf{F_V}$.

\vspace{2mm}    
\noindent \textbf{Surface-aligned Features}
    Although they capture a global context about the shape of a 3D object, voxel-aligned features are queried from a low-resolution volumetric grid and thus lack geometric information on surface.
    As a complement, we introduce surface-aligned features that capture fine details of surface to facilitate radiance field learning.  
    Given an input image $\mathbf{I}$, we first regress a sparse point cloud $\mathcal{S}$ of size 1024 from $\mathbf{I}$ using a point set generator $G_\mathcal{S}$ based on GraphX-convolutions~\cite{nguyen2019graphx}. 
    Then, we feed the generated point cloud to a PointNet++~\cite{qi2017pointnet++} network to extract point-wise features $\mathbf{F}_\mathcal{S}$. 
    For each query point $\mathbf{X}$, we define its surface-aligned feature $\mathbf{f}_\mathcal{S}$ as the weighted sum of the corresponding feature vectors of $\mathbf{X}$'s $K$-nearest neighbors in $\mathcal{S}$:
    \begin{equation}
        \mathbf{f}_\mathcal{S} = \sum_{k=0}^{K} w_k * \mathbf{F}_{\mathcal{S}_{m(k)}}
        \label{surfalign}
    \end{equation}
    where $m(k), k=0,1,2...,K$ is the indices of the $K$ points, $w_k$ is inversely proportional to its distance to $\mathbf{X}$:
    \begin{equation}
        w_k = 1/(1+ \exp(||\mathbf{X}-\mathcal{S}_{m(k)}||) 
        \label{weight}
    \end{equation}
    In this way, the features from the nearest neighbor contributes most to the $\mathbf{f}_\mathcal{S}$, and vice versa.

\subsection{Radiance Field Prediction and Rendering}
    Given an input single-view image $\mathbf{I}$, we construct its radiance field with a MLP $f$, which regresses the volume density $\mathbf{\sigma}$ and view-dependent radiance $\mathbf{r}$ from the 3D coordinates of a query point $\mathbf{X}$, a viewing direction $\mathbf{d}$, and the corresponding pixel-, voxel-, and surface-aligned features (~\ie $\mathbf{f_I}$, $\mathbf{f_V}$, and $\mathbf{f_S}$) of $\mathbf{I}$:
    \begin{equation}
        \sigma, \mathbf{r} = f(\gamma_m(\mathbf{X}), \gamma_n(\mathbf{d}); \mathbf{f_I} \oplus \mathbf{f_V} \oplus \mathbf{f_S})
    \label{MLP}
    \end{equation}
    where $\gamma_m$ and $\gamma_n$ are
    position encoding functions~\cite{vaswani2017attention, mildenhall2020nerf} applied to $\mathbf{X}$,  $\mathbf{d}$ respectively, which alleviates the positional bias inherent in Cartesian coordinates without sacrificing their discrepancy in-between. Specifically, $\gamma$ maps Cartesian coordinates from $\mathbb{R}$ into a high dimensional space $\mathbb{R}^{2L}$:
    \begin{equation}
    \begin{aligned}
        \gamma_L(\mathbf{p}) = ( \sin(2^0 \pi \mathbf{p}), &\cos(2^0 \pi \mathbf{p}), \\ &...,
        \sin(2^{L-1} \pi \mathbf{p}), \cos(2^{L-1} \pi \mathbf{p}) )
    \end{aligned}
    \end{equation}
    where $\gamma(\cdot)$ is applied separately to each component of vector $\mathbf{p}$.
    With the constructed radiance field represented by $\sigma$ and $\mathbf{r}$, we render novel view images via differentiable rendering as:
    \begin{equation}
        \mathbf{c_t} = \sum_i \tau_i (1 - \exp(-\sigma_i)) \mathbf{r_i} 
        \label{VolRendering}
    \end{equation}
    where $\mathbf{c_t}$ is the rendered pixel color, $\tau_i = \exp (- \sum_{j=1}^{i-1} \sigma_j)$ denotes the volume transmittance.
    Intuitively, Eq.~\ref{VolRendering} shows that the a pixel's radiance value (\ie RGB color) can be calculated by casting a ray from the camera to the pixel and accumulating the radiance of points sampled on the ray.

 \subsection{Loss Functions}
    
    Corresponding to our pixel-, voxel- and surface-aligned features, we train our model using three different loss functions as follows.
    
    \vspace{2mm}
    \noindent \textbf{RGB Rendering Loss}
    Similar to existing works in the NeRF series, we use $L2$ rendering loss as the main loss function. It constrains that the rendered color value of each ray should be consistent with the corresponding ground-truth pixel value. Thus, we have:
    \begin{equation}
        L_r = || \mathbf{c_t} - \mathbf{\hat{c_t}} ||_2^2
        \label{rendeingloss}
    \end{equation}
    where $\mathbf{c_t}$ and $\mathbf{\hat{c_t}}$ are the predicted and ground-truth color values of sampled pixels from novel view $\mathbf{I_t}$ with viewpoint $\mathbf{R_t}$ respectively.
    
    \vspace{2mm}
    \noindent \textbf{Volume Reconstruction Loss}
    To learn volumetric features $\mathbf{F_V}$, we add a 3D convolutional layer after $\mathbf{F_V}$ to estimate a low-resolution occupancy volume $
    \mathbf{V} \in \mathbb{R}^{32 \times 32 \times 32}$, whose ground-truth label is $\mathbf{V}^{*}$. Then, we apply a standard binary cross-entropy loss and have:
    \begin{equation}
        L_v = \sum_{i \in [1:32]^3 } \mathbf{V}^{*}(i) \log \mathbf{V}(i)  +
        (1-\mathbf{V}^{*}(i)) \log (1-\mathbf{V}(i))
        \label{bceloss}
    \end{equation}
    
    \vspace{2mm}
    \noindent \textbf{Point Regression Loss}
    We employ the Chamfer distance to constraint our point set generation and have:
    \begin{equation}
        L_p = \sum_{\mathbf{q} \in \mathcal{S}} \min_{\mathbf{q}^{*} \in \mathcal{S}^{*}} \| \mathbf{q} - \mathbf{q}^{*} \|^2 \ \ + 
        \sum_{\mathbf{q}^{*} \in \mathcal{S}^{*}} \min_{\mathbf{q} \in \mathcal{S}} \| \mathbf{q} - \mathbf{q}^{*} \|^2
    \label{chamfer}
    \end{equation}
    where $\mathcal{S}$ is the predicted point set and $\mathcal{S}^*$ is its corresponding ground truth.
    
    \vspace{2mm}
    \noindent \textbf{Overall Loss Function}
    Our overall loss function is:
    \begin{equation}
        L = \lambda_1 * L_r + \lambda_2 * L_v + \lambda_3 * L_p
        \label{totaloss}
    \end{equation}
    where $\lambda_1$, $\lambda_2$, and $\lambda_3$ are weighting parameters.

\section{Experiments}
\label{sec:experiments}

\definecolor{myyellow}{rgb}{1,1, 0.6}
\definecolor{myorange}{rgb}{1, 0.8, 0.6}
\definecolor{myred}{rgb}{1, 0.6, 0.6}
\newcommand{\tablefirst}[0]{\cellcolor{myred}}
\newcommand{\tablesecond}[0]{\cellcolor{myorange}}
\newcommand{\tablethird}[0]{\cellcolor{myyellow}}

\begin{table*}[!t]
\resizebox{\textwidth}{!}{%
\begin{tabular}{@{}ll||llllllllllllll@{}}
\toprule
 & \multicolumn{1}{l}{} & plane & bench & cbnt. & car & chair & disp. & lamp & spkr. & rifle & sofa & table & phone & boat & mean \\ 
 
 \midrule
 & DVR~\cite{niemeyer2020differentiable} & 25.29 & 22.64 & 24.47 & 23.95 & 19.91 & 20.86 & 23.27 & 20.78 & 23.44 & 23.35 & 21.53 & 24.18 & 25.09 & 22.70 \\
$\uparrow$ PSNR & SRN~\cite{sitzmann2019scene} & 26.62 & 22.20 & 23.42 & 24.40 & 21.85 & 19.07 & 22.17 & 21.04 & 24.95 & 23.65 & 22.45 & 20.87 & 25.86 & 23.28 \\
 & pixelNeRF~\cite{yu2021pixelnerf} & \tablesecond29.76 & \tablesecond26.35 & \tablesecond27.72 & \tablesecond27.58 & \tablesecond23.84 & \tablesecond24.22 & \tablesecond28.58 & \tablesecond24.44 & \tablesecond30.60 & \tablesecond26.94 & \tablesecond25.59 & \tablefirst27.13 & \tablesecond29.18 & \tablesecond26.80 \\ 
 & Ours & \tablefirst31.32 & \tablefirst27.43 & \tablefirst28.40 & \tablefirst28.12 & \tablefirst24.37 & \tablefirst24.61 & \tablefirst28.73 & \tablefirst24.44 & \tablefirst30.82 & \tablefirst27.42 & \tablefirst26.60 & \tablesecond26.99 & \tablefirst29.92 & \tablefirst27.48 \\ 
 
 \midrule
 & DVR~\cite{niemeyer2020differentiable} & 0.905 & 0.866 & 0.877 & 0.909 & 0.787 & 0.814 & 0.849 & 0.798 & 0.916 & 0.868 & 0.840 & 0.892 & 0.902 & 0.860 \\
$\uparrow$ SSIM & SRN~\cite{sitzmann2019scene} & 0.901 & 0.837 & 0.831 & 0.897 & 0.814 & 0.744 & 0.801 & 0.779 & 0.913 & 0.851 & 0.828 & 0.811 & 0.898 & 0.849 \\
 & pixelNeRF~\cite{yu2021pixelnerf} & \tablesecond0.947 & \tablesecond0.911 & \tablesecond0.910 & \tablefirst0.942 & \tablesecond0.858 & \tablesecond0.867 & \tablesecond0.913 & \tablefirst0.855 & \tablefirst0.968 & \tablesecond0.908 & \tablesecond0.898 & \tablefirst0.922 & \tablesecond0.939 & \tablesecond0.910 \\ 
 & Ours & \tablefirst0.956 & \tablefirst0.923 & \tablefirst0.912 & \tablesecond0.940 & \tablefirst0.869 & \tablefirst0.867 & \tablefirst0.915 & \tablesecond0.853 & \tablesecond0.965 & \tablefirst0.912 & \tablefirst0.911 & \tablesecond0.915 & \tablefirst0.940 & \tablefirst0.915 \\ 
 
 \midrule
 & DVR~\cite{niemeyer2020differentiable} & 0.095 & 0.129 & 0.125 & 0.098 & 0.173 & 0.150 & 0.172 & 0.170 & 0.094 & 0.119 & 0.139 & 0.110 & 0.116 & 0.130 \\
$\downarrow$ LPIPS & SRN~\cite{sitzmann2019scene} & 0.111 & 0.150 & 0.147 & 0.115 & 0.152 & 0.197 & 0.210 & 0.178 & 0.111 & 0.129 & 0.135 & 0.165 & 0.134 & 0.139 \\
 & pixelNeRF~\cite{yu2021pixelnerf} & \tablesecond0.084 & \tablesecond0.116 & \tablesecond0.105 & \tablesecond0.095 & \tablesecond0.146 & \tablefirst0.129 & \tablesecond0.114 & \tablesecond0.141 & \tablesecond0.066 & \tablesecond0.116 & \tablesecond0.098 & \tablefirst0.097 & \tablesecond0.111 & \tablesecond0.108 \\ 
 & Ours & \tablefirst0.065 & \tablefirst0.098 & \tablefirst0.097 & \tablefirst0.087 & \tablefirst0.128 & \tablesecond0.133 & \tablefirst0.104 & \tablefirst0.140 & \tablefirst0.066 & \tablefirst0.104 & \tablefirst0.082 & \tablesecond0.107 & \tablefirst0.101 & \tablefirst0.096 \\ 
 
 \bottomrule
\end{tabular}%
}
	\caption{\textbf{Quantitative comparison on category-agnostic view synthesis.} We color code each row as \colorbox{myred}{\textbf{best}} and \colorbox{myorange}{\textbf{second best}}. Our method outperforms all baselines by a wide margin in terms of all mean metrics.}
	\label{tab:agnostic}
	\vspace{-3mm}
\end{table*}
\begin{figure*}[!t]
	\centering
	\includegraphics[width=1.\linewidth]
    {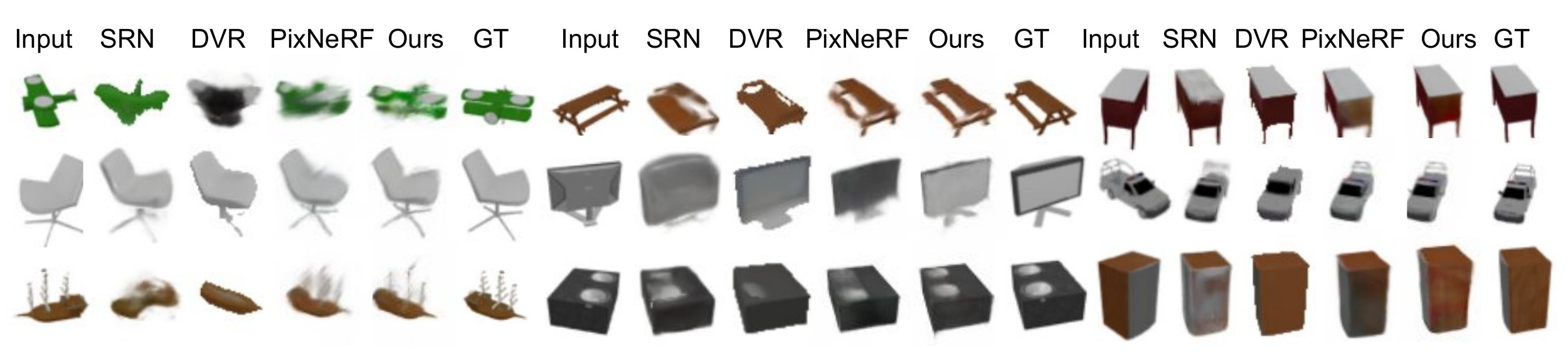}
	\caption{\textbf{Qualitative comparison on category-agnostic view synthesis.} A single model is trained among 13 ShapeNet categories, and tested on a single image for novel view synthesis. We observe that our method produces detailed novel views, and is consistent in both geometry and appearance. Conversely, pixelNeRF~\cite{yu2021pixelnerf} fails to infer correct geometry and produce inconsistent and blurry textures.}
	\label{fig:agnostic}
	\vspace{-5mm}
\end{figure*}

To demonstrate the superiority of our PVSeRF, we first compare it against state-of-the-art methods on two single-image novel view synthesis tasks, \ie category-agnostic view synthesis and category-specific view synthesis. Then, we evaluate our approach on real images, demonstrating the generalization ability of our method. Finally, we conduct ablation studies to validate the  effectiveness of each component of our PVSeRF. \\

\mypar{Datasets}
\label{dataset}
We benchmark our method extensively on the synthetic images from the ShapeNet~\cite{chang2015shapenet} dataset. Specifically, for the category-agnostic view synthesis task, we use the renderings and splits from Kato \etal~\cite{kato2018neural} which renders objects from 13 categories of the ShapeNetCore-V1 dataset. Each object was rendered at 64$\times$64 resolution from 24 equidistant azimuth angles, with a fixed elevation angle. For the category-specific view synthesis task, we use the dataset and splits provided by Sitzmann \etal~\cite{sitzmann2019scene}, which renders 6,591 chairs and 3,514 cars from the ShapeNetCore V2 dataset. For the evaluation on real images, we use the collected real-world cars images from~\cite{KrauseStarkDengFei-Fei_3DRR2013}. To provide supervision for volume reconstruction and point set regression, we convert each ground-truth mesh to a point set of size $2048$ and a volumetric grid of resolution $32^3$.

\paragraph{Implementation Details}
 We implement our model with PyTorch~\cite{NEURIPS2019_9015}.
 Details of the network architecture are presented in the supplementary material.
 The training process of our approach consists of two stages:
 i) we pre-train the volume generator $ G_{\mathbf{V}}$ and the point set generator $ G_{\mathbf{S}}$ respectively using loss functions defined in Eq.~\ref{bceloss} and Eq.~\ref{chamfer}. 
 Specifically, $ G_{\mathbf{V}}$ is trained with an initial learning rate of $10^{-3}$ and a batch size of $64$ for $250$ epochs. The learning rate drops by a factor of $5$ after $150$ epochs. 
 $ G_{\mathcal{S}}$ is trained with an initial learning rate of $10^{-5}$ and a batch size of $4$ for $10$ epochs. The learning rate drops by a factor of $3$ after $5$ and $8$ epochs. 
 ii) we fine-tune the whole network for $400$ epochs. We set the learning rate as $10^{-4}$ and the batch size as $4$. We use an Adam~\cite{kingma2014adam} optimizer for all the training mentioned above.
 We empirically set hyper-parameters $s=0.0722$, $K=5$, $m=6$, $n=0$, $\lambda_1=\lambda_2=\lambda_3=1$.

\paragraph{Evaluation Protocol} 
Following the community standards \cite{mildenhall2020nerf, sitzmann2019scene, niemeyer2020differentiable}, we use peak signal-to-noise ratio (PSNR) and structural similarity index (SSIM)~\cite{wang2004image} to measure the quality of the synthesized novel views. 
We also use LPIPS~\cite{zhang2018unreasonable} that has been shown to be closer to human perception.

\subsection{Category-agnostic View Synthesis}

Category-agnostic novel view synthesis aims to learn object priors that can generalize across multiple categories.
\vspace{-3mm}

\paragraph{Baselines}
We compare our method against three closely-related state-of-the-art methods: SRN~\cite{sitzmann2019scene}, DVR~\cite{niemeyer2020differentiable} and pixelNeRF~\cite{yu2021pixelnerf}, which are applicable to synthesize novel views for all categories. For DVR and pixelNeRF, we use pretrained models from their official Github repositories\footnote{Niemeyer~\etal~\cite{niemeyer2020differentiable}: \href{https://github.com/autonomousvision/differentiable_volumetric_rendering}{https://github.com/autonomousvision/differentiable\\\_volumetric\_rendering}, Yu~\etal~\cite{yu2021pixelnerf}: \href{https://github.com/sxyu/pixel-nerf}{https://github.com/sxyu/pixel-nerf}.}. For SRN~\cite{sitzmann2019scene}, we use the model trained by~\cite{yu2021pixelnerf} to make it comparable with~\cite{sitzmann2019scene, yu2021pixelnerf}. All methods are trained using the same dataset and settings introduced in~\cref{dataset}.
To facilitate a fair comparison, we follow the random view indices provided by pixelNeRF and select the input view for each test object accordingly.

\vspace{-3mm}
\paragraph{Results}
As~\cref{fig:agnostic} shows, our method outperforms all previous methods by synthesizing more detailed novel views. 
In addition, it can be observed that i) the two baseline methods, DVR~\cite{niemeyer2020differentiable} and SRN~\cite{sitzmann2019scene}, tend to generate blurry images and distorted geometries; ii) pixelNeRF~\cite{yu2021pixelnerf} shows blurry and inconsistent appearance.
The quantitative results in \Cref{tab:agnostic} further justify the superiority of our method against all baselines in terms of the mean values of PSNR, SSIM and LPIPS metrics.
Notably, the PSNR of our approach attains a significant improvement over the second best method by $0.68$.

\subsection{Category-specific View Synthesis}
\label{sec:specific}

For category-specific view synthesis, all methods are trained on the chair or car categories of ShapeNet~\cite{chang2015shapenet}.
\vspace{-3mm}

\paragraph{Baselines}
We choose SRN~\cite{sitzmann2019deepvoxels} and pixelNeRF~\cite{yu2021pixelnerf} as the baseline methods. We also report the quantitative results from TCO~\cite{Tatarchenko2015SingleviewTM} and dGQN~\cite{eslami2018neural} provided by~\cite{sitzmann2019scene}, to keep in line with prior arts.

\begin{figure}[!t]
	\centering
	\includegraphics[width=1.\linewidth]{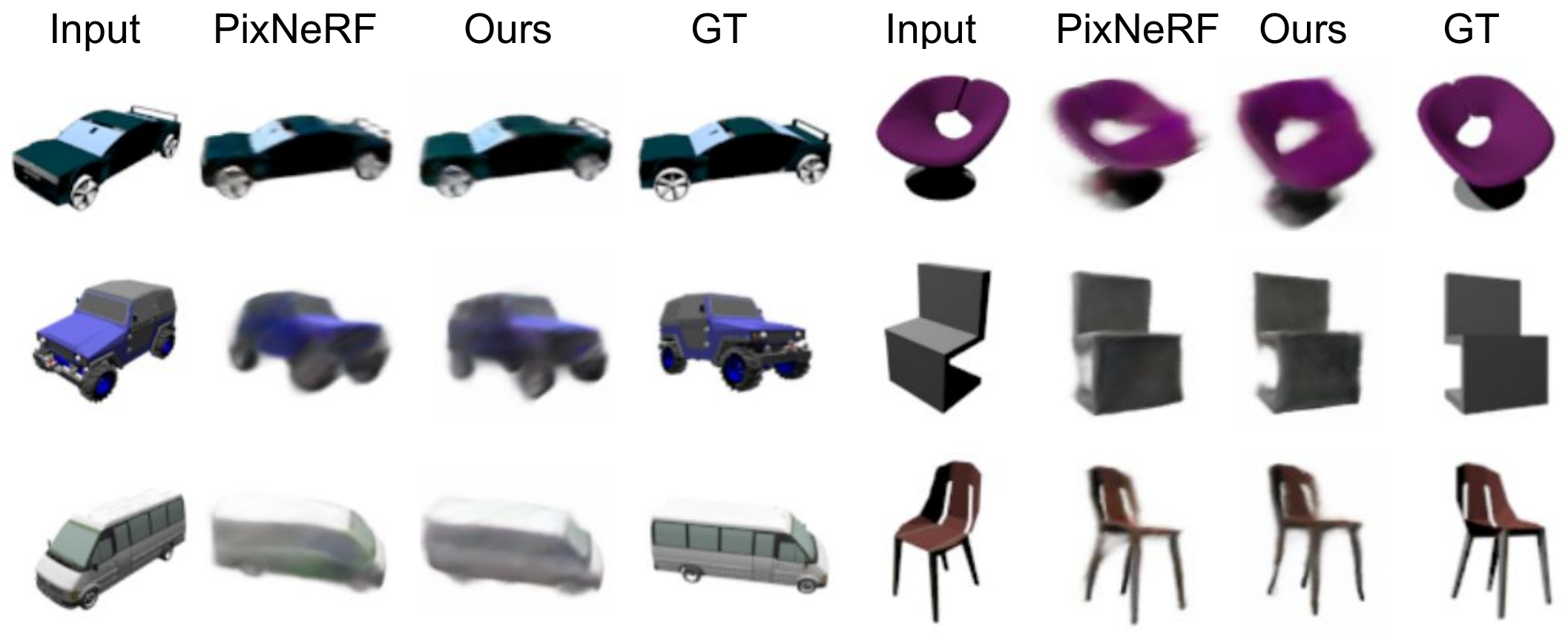}
%	\includegraphics[width=1.\linewidth]{figures/teaser}
% 	\caption{\textbf{Qualitative comparison on category-specific view synthesis.} Although our overall quantitative evaluation is slightly worse than pixelNeRF~\cite{yu2021pixelnerf}, the visual results which have normal input views still remain comparable to the strong baselines.}
\caption{\textbf{Qualitative comparison on category-specific view synthesis.} The performance of our method is comparable to that of the state-of-the-art pixelNeRF~\cite{yu2021pixelnerf}.}
	\label{fig:specific}
%	\vspace{-2mm}
\end{figure}
\begin{figure}[t]
	\centering
	\includegraphics[width=1.\linewidth]{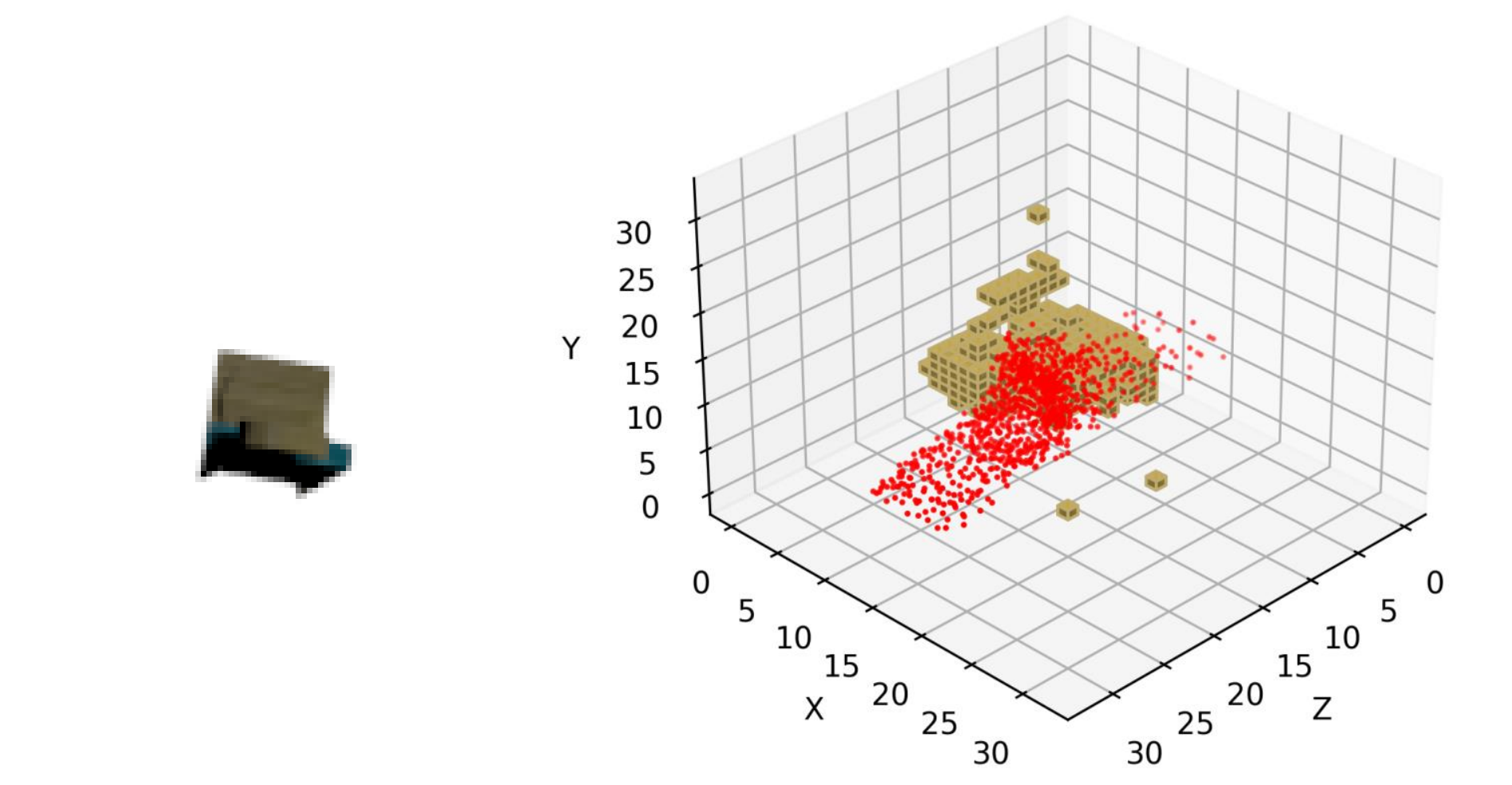}
	\caption{\textbf{Failure case of explicit geometry reasoning.} Under the challenging viewpoint, the scene geometry is ambiguously captured in a single image, causing the network being unable to predict plausible geometries.}
	\label{fig:failure_geometry}
	\vspace{-5mm}
\end{figure}

\vspace{-3mm}
\paragraph{Results}

We show the quantitative and qualitative results in~\Cref{tab:specific} and~\cref{fig:specific} respectively. 
It can be observed that the performance of our method is comparable to the state-of-the-art method~\cite{yu2021pixelnerf} both qualitatively and quantitatively.
Such comparable results indicate that the advantages of our method are not significant in some neural rendering cases. 
We carefully investigate the results and ascribe this to the invalidity of explicit geometry reasoning in some cases (\cref{fig:failure_geometry}). Since the renderings provided by~\cite{sitzmann2019scene} contain many challenging camera viewpoints, the explicit geometry reasoning from single-view becomes a more challenging problem. We postpone the discussion of this phenomenon to~\cref{sec:conclusion}.

\begin{figure*}[htb]
	\centering
	\includegraphics[width=1.\linewidth]{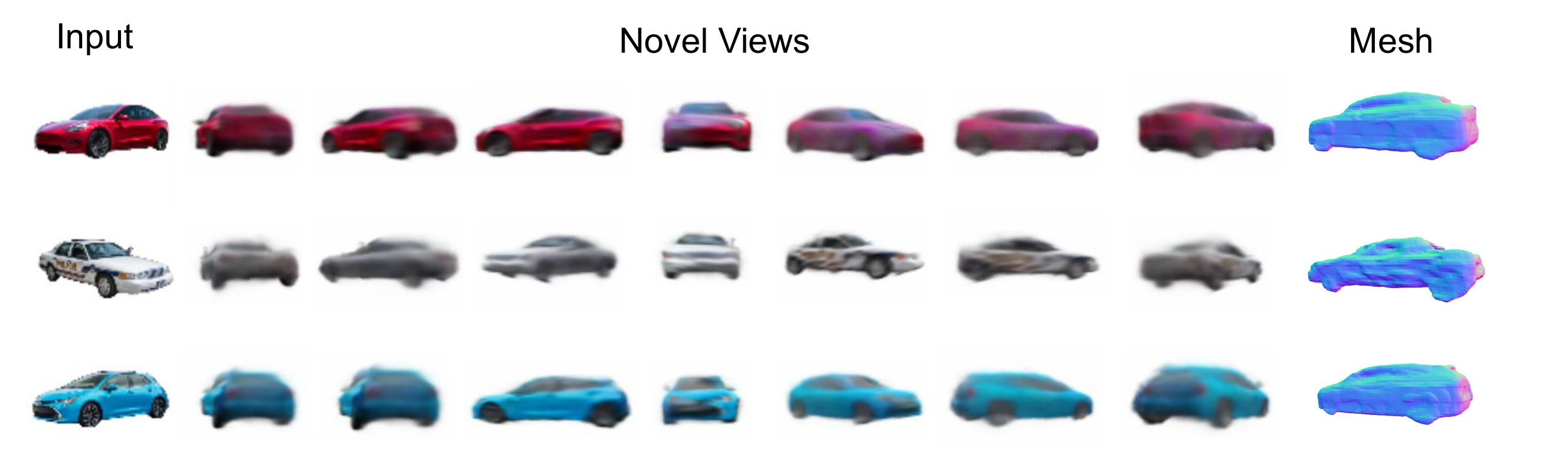}
	\caption{\textbf{Novel view synthesis results on real car images.} Although extensively trained on synthetic data, our method can easily generalize to real single-view images, and produce plausible view synthesis results and underlying geometries.}
	\label{fig:real}
% 	\vspace{-2mm}
\end{figure*}

\subsection{Novel View Synthesis on Real Images}
 To highlight the generalization ability of our method, we evaluate our pretrained models directly on real images without any finetuning. 
 Specifically, we first take the images from the Stanford cars dataset~\cite{KrauseStarkDengFei-Fei_3DRR2013} and apply the PointRend model~\cite{Kirillov2020PointRendIS} to mask their clutter backgrounds. Then, we feed the preprocessed images into a category-specific model of ShapeNet ``cars'' to predict novel views. As Fig.~\ref{fig:real} shows, our method can not only synthesize visually compelling novel views, but also infer accurate geometries. This effectively demonstrates the excellent generalization performance of our method on real image as it is only trained on synthetic images.
\begin{table}[!t]
	\centering
    \resizebox{.9\linewidth}{!}{ % <atuo-adjusts font size to fill
    \begin{tabular}{@{}llll@{}}
    \toprule
    %& \multicolumn{1}{l}{} & \multicolumn{2}{c}{1-view} \\  \cmidrule(lr){3-4}
  %  & \multicolumn{1}{l}{} & PSNR & SSIM \\ \midrule[0.8]
   & \multicolumn{1}{l}{} & PSNR$\uparrow$ & SSIM$\uparrow$ \\  \midrule
    & TCO~\cite{Tatarchenko2015SingleviewTM} & 21.27 & 0.88 \\
    & dGQN~\cite{eslami2018neural} & 21.59 & 0.87 \\
    Chairs & SRN~\cite{sitzmann2019scene} & 22.89 & 0.89 \\
    & pixelNeRF~\cite{yu2021pixelnerf} & \textbf{23.72} & \textbf{0.91} \\
    & Ours & 23.33 & \textbf{0.91} \\ \midrule
    & SRN~\cite{sitzmann2019scene} & 22.25 & 0.89 \\ 
    Cars & pixelNeRF~\cite{yu2021pixelnerf} & \textbf{23.17} & \textbf{0.90} \\
    & Ours & 22.98 & \textbf{0.90} \\ \bottomrule
    \end{tabular}
    }%<\resizebox
    \caption{\textbf{Quantitative comparison on category-specific view synthesis.} Since the renderings from~\cite{sitzmann2019scene} contain many challenging camera viewpoints, our performance is degenerated ascribe to the invalidity of explicit geometry reasoning. Nevertheless, our method is comparable to the state-of-the-art method~\cite{yu2021pixelnerf}.}
    \label{tab:specific}
    
\end{table}

\begin{figure}[t]
	\centering
	\includegraphics[width=1.\linewidth]{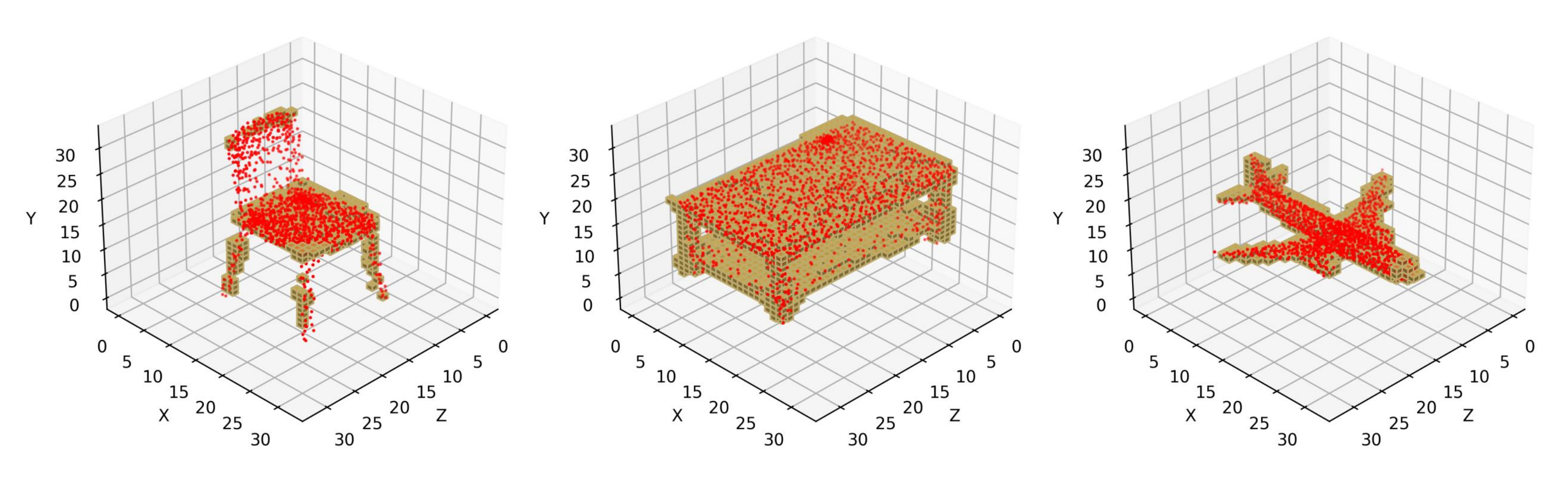}
	\caption{\textbf{Illustration of the complementary properties of point set and volumes.} We randomly show several predicted geometries. It can be seen that these two representations exhibit reciprocal behaviors: the missing parts of volumetric grid are spanned by point set, while the regions where the point set is too sparse are occupied by volumes.}
	\label{fig:geometry}
	\vspace{-5mm}
\end{figure}
\begin{table*}[]
	\resizebox{\textwidth}{!}{%
		\begin{tabular}{@{}ll||llllllllllllll@{}}
			\toprule
			& \multicolumn{1}{l}{} & plane & bench & cbnt. & car & chair & disp. & lamp & spkr. & rifle & sofa & table & phone & boat & mean \\ 
			
			\midrule
			& w/o joint & 29.03 & 25.18 & 26.12 &  25.82 & 21.97 & 22.25 & 26.33 & 22.19 & 28.55 & 25.18 & 24.27 & 24.67 & 27.54 & 25.16 \\ 
			& w/o surface-aligned  & 30.82 & 27.00 & 28.31 & 27.67 & 24.05 & 24.33 & 28.73 & 24.33 & 30.63 & 26.97 & 26.27 & 26.85 & 29.58 & 27.15 \\
			$\uparrow$ PSNR & w/o voxel-aligned & 30.83 & 27.14 & \textbf{28.40} & 27.93 & 24.35 & \textbf{24.66} & \textbf{29.10} & \textbf{24.75} & \textbf{31.05} & 27.29 & 26.48 & \textbf{27.01} & 29.61 & 27.38 \\
			& Ours & \textbf{31.32} & \textbf{27.43} & \textbf{28.40} & \textbf{28.12} & \textbf{24.37} & 24.61 & 28.73 & 24.44 & 30.82 & \textbf{27.42} & \textbf{26.60} & 26.99 & \textbf{29.92} & \textbf{27.48} \\ 
			
			\midrule
			& w/o joint & 0.926 & 0.863 & 0.864 & 0.924 & 0.844 & 0.803 & 0.826 & 0.812 & 0.947 & 0.878 & 0.853 & 0.836 & 0.923 & 0.876 \\ 
			& w/o surface-aligned & 0.954 & 0.917 & 0.912 & 0.936 & 0.860 & 0.860 & \textbf{0.915} & 0.849 & \textbf{0.967} & 0.904 & 0.906 & 0.912 & \textbf{0.940} & 0.911 \\
			$\uparrow$ SSIM & w/o voxel-aligned & 0.955 & 0.921 & \textbf{0.913} & \textbf{0.941} & 0.867 & \textbf{0.870} & \textbf{0.915} & \textbf{0.856} & 0.966 & 0.911 & 0.910 & \textbf{0.917} & 0.939 & \textbf{0.915} \\
			& Ours & \textbf{0.956} & \textbf{0.923} & 0.912 & 0.940 & \textbf{0.869} & 0.867 & \textbf{0.915} & 0.853 & 0.965 & \textbf{0.912} & \textbf{0.911} & 0.915 & \textbf{0.940} & \textbf{0.915} \\ 
			
			\midrule
			& w/o joint & 0.097 & 0.134 & 0.134 & 0.099 & 0.137 & 0.163 & 0.175 & 0.164 & 0.098 & 0.123 & 0.120 & 0.149 & 0.120 & 0.123 \\ 
			& w/o surface-aligned & 0.064 & 0.100 & \textbf{0.096} & 0.094 & 0.136 & 0.132 & 0.102 & 0.144 & 0.064 & 0.108 & 0.085 & \textbf{0.100} & \textbf{0.100} & 0.099 \\
			$\downarrow$ LPIPS & w/o voxel-aligned & \textbf{0.063} & \textbf{0.096} & \textbf{0.096} & \textbf{0.085} & 0.130 & \textbf{0.126} & \textbf{0.100} & \textbf{0.137} & \textbf{0.060} & \textbf{0.104} & \textbf{0.081} & 0.104 & \textbf{0.100} & \textbf{0.095} \\
			& Ours & 0.065 & 0.098 & 0.097 & 0.087 & \textbf{0.128} & 0.133 & 0.104 & 0.140 & 0.066 & \textbf{0.104} & 0.082 & 0.107 & 0.101 & 0.096 \\ 
			
			\bottomrule
		\end{tabular}%
	}
	\caption{\textbf{Quantitative comparison of ablation studies.} Our joint method that employs complementary coarse volumetric features and fine surface features achieves the best performance. Whereas, removing any part of the proposed method will cause more or less deterioration.}
	\label{tab:ablation}
	%\vspace{-3mm}
\end{table*}

\subsection{Ablation Study}
\label{ablation}

To validate the effectiveness of each proposed component, we conduct an ablation study on our method, yielding three variants: i) {\it w/o surface-aligned feature}, in which only pixel- and voxel-aligned features are incorporated; ii) {\it w/o voxel-aligned feature}, where the radiance field is conditioned only on pixel- and surface-aligned features; iii) {\it w/o joint training}, in which we fix all feature extractors\footnote{We use a PointNet++~\cite{qi2017pointnet++} model trained on PartNet~\cite{Mo_2019_CVPR} segmentation task as a point-feature extractor.} and solely train the radiance field predictor $f$.
As~\Cref{tab:ablation} shows, it can be observed that the {\it w/o joint training} variant constantly performs the worst among all variants. This demonstrates that the joint learning of pixel-, voxel- and surface-aligned features is crucial in our explicit geometric reasoning. 
In addition, the performance of the {\it w/o surface-aligned} variant is always worse than the {\it w/o voxel-aligned feature} variant, as the voxel-aligned features queried from a low-resolution volume are better at capturing global geometry contexts.
Our full method achieves the best performance among all variants, which validates the effectiveness of employing a hybrid of geometric features that complement each other. This is also demonstrated in \cref{fig:geometry}.

% \section{Conclusion}
% \label{sec:conclusion}

\section{Conclusion}
\label{sec:conclusion}
For the task of novel view synthesis from single-view RGB images, we present PVSeRF, a novel learning framework that reconstructs neural radiance fields conditioned on joint pixel-, voxel-, and surface-aligned features. By augmenting hybrid geometric features with image features, we effectively address the feature confusion issue of pixel-aligned features. Compare to previous arts, our framework gains superior or comparable results in terms of both visual perception and quantitative measures. Moreover, a suite of ablation studies also verify the efficacy of our key contributions.

\vspace{2mm}
\paragraph{Limitation and Future Works}
Despite the effectiveness of our method, there are still some limitations to be addressed in future work.
First, the performance of our method is dependent on the amount of geometric information within the input single-view image. As discussed in~\cref{sec:specific}, when the scene geometry is little captured in the input image due to challenging viewpoints, the novel views synthesized by our method may become less clear. 
In future work, we plan to include multi-view consistency as an additional supervision to train our geometry reasoning network, thereby increasing its robustness to challenging viewpoints.
Secondly, we focus on the geometry reasoning from complete geometries (\ie surface and voxel) of 3D shapes  that reconstructs neural radiance fields from single-view RGB image and have not investigated that from more challenging partial geometries (\eg depth maps or multiplane images~\cite{tucker2020single, wizadwongsa2021nex}). 
In future work, we plan to extend our method to such partial geometries, thereby making our method more flexible.

%%%%%%%%% REFERENCES
{\small
\bibliographystyle{ieee_fullname}
\bibliography{PVSeRF}
}

\end{document}